%% file: nips_2017.tex
\documentclass{article}

\input{packages.tex}
\title{Causality Refined Diagnostic Prediction}
\author{
  Marcus Klasson$^{*}$, Kun Zhang$^{\dagger}$, Bo C. Bertilson$^{\ddagger}$, Cheng Zhang$^{*}$, Hedvig Kjellstr\"{o}m$^{*}$ \\
$^{*}$KTH Royal Institute of Technology \hspace{1em} $^{\dagger}$Carnegie Mellon University \hspace{1em} $^{\ddagger}$Karolinska Institute \\
\texttt{\{mklas, chengz, hedvig\}@kth.se} \hspace{3em} \texttt{kunz1@cmu.edu}  \hspace{2.5em} \texttt{bo.bertilson@ki.se} \\
}

\begin{document}

\maketitle

\begin{abstract}
Applying machine learning in the health care domain has shown promising results in recent years. Interpretable outputs from learning algorithms are desirable for decision making by health care personnel. In this work, we explore the possibility of utilizing causal relationships to refine diagnostic prediction.  We focus on the task of diagnostic prediction using discomfort drawings, and explore two ways to employ causal identification to improve the diagnostic results. Firstly, we use causal identification to infer the causal relationships among diagnostic labels which, by itself, provides interpretable results to aid the decision making and training of health care personnel. Secondly, we suggest a post-processing approach where the inferred causal relationships are used to refine the prediction accuracy of a multi-view probabilistic model. Experimental results show firstly that causal identification is capable of detecting the causal relationships among diagnostic labels correctly, and secondly that there is potential for improving pain diagnostics prediction accuracy using the causal relationships.  
\end{abstract}

\input{intro.tex}
\input{model.tex}

\input{exp.tex}

\input{discussion.tex}

\bibliographystyle{plain}
\bibliography{ref}

\end{document}

%% file: packages.tex
\usepackage[final, nonatbib]{nips_2017}

\usepackage[utf8]{inputenc} 
\usepackage[T1]{fontenc}    
\usepackage{url}            
\usepackage{booktabs}       
\usepackage{amsfonts}       
\usepackage{nicefrac}       
\usepackage{microtype}      

\usepackage{authblk}
\usepackage{color}
\usepackage{algorithm}
\usepackage{algpseudocode}
\usepackage{graphicx}
\usepackage{epstopdf}
\usepackage{enumitem}
\usepackage{amssymb}
\usepackage{amsmath}

\usepackage{color,comment,rotating,subfigure,url,xspace} 
\usepackage{tikz}
\usetikzlibrary{arrows,shadows,shapes,backgrounds,decorations,snakes,fit}
\usepackage{todonotes}
\usepackage{enumerate}
\usepackage{footnote}
\usepackage{lipsum}

\usepackage{titlesec}
\titlespacing\section{0pt}{2pt plus 2pt minus 2pt}{2pt plus 2pt minus 2pt}
\titlespacing\subsection{0pt}{2pt plus 2pt minus 2pt}{2pt plus 2pt minus 2pt}
\titlespacing\subsubsection{0pt}{0pt plus2pt minus 2pt}{0pt plus 2pt minus 2pt}
\titlespacing\paragraph{0pt}{0pt plus 2pt minus 2pt}{0pt plus 2pt minus 2pt}

%% file: intro.tex
\section{Introduction}

\begin{figure}[t]
\begin{minipage}{0.25\textwidth}
\includegraphics[height=4cm, width=\textwidth]{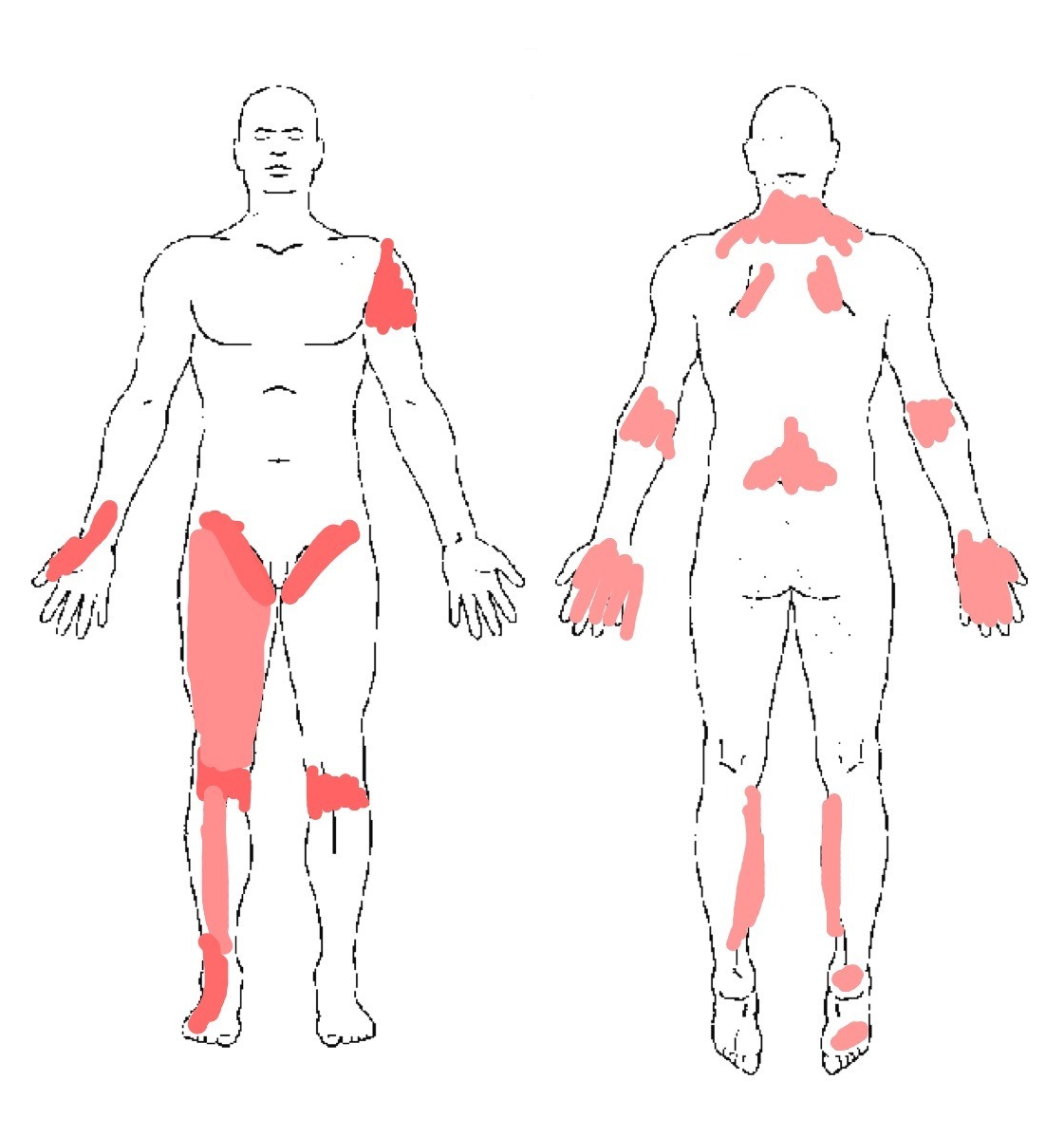} 
\end{minipage}%
\begin{minipage}{0.70\textwidth}
{\small
\begin{itemize}
\setlength{\itemsep}{-4pt}
\item[] \textbf{Symptom diagnoses:} Neck discomfort, B Scapula discomfort, R Shoulder impingement, Interscapular discomfort, B Medial elbow discomfort, Lumbago, R Thumb discomfort, B Adductor tendonitis, R Front thigh discomfort, B PFS, B Calf discomfort, R Shin discomfort, R Calcaneal pain, R Arch discomfort\\
\item[] \textbf{Pattern diagnoses:} B C7 Radiculopathy, B L1 Radiculopathy, B L5 Radiculopathy, R S1 Radiculopathy, R C6 Radiculopathy \\
\item[] \textbf{Psychophysiological diagnoses:} DLI C6-C7, DLI L4-L5, DLI L5-S1, DLI S1-S2, DLI S2-S3
\end{itemize}}
\end{minipage}
\caption{\footnotesize A discomfort drawing (left) and assessment from a medical expert (right). R stands for right side, L for left side and B for bilateral. PFS refers to patellofemoral pain syndrome and DLI to disco-ligament injury.}
\label{fig:discomfort_drawing}
\end{figure}

Using pain or discomfort drawings for diagnostic purposes has shown to be effective both in medical science \cite{bertilson2003reliability,bertilson2007pain,bertilson2010assessment} and machine learning \cite{zhang2016diagpred}. Pain drawings have a long history in the health care system and the first research paper on the use of it was published in 1949 \cite{palmer1949pain}.  It is used in many clinics and especially in the assessment of spine-related pain \cite{ohnmeiss1999relation, vucetic1995pain, albeck1996critical, tanaka2006cervical}. Figure \ref{fig:discomfort_drawing} shows an example of a discomfort drawing (a modern version of the original pain drawing) and its assessed diagnostic labels. The drawing is filled out by the patient who shadows those parts of the body where he/she experiences discomfort. Health care personnel then assess the drawing to decide on diagnosis or further examinations. The drawing has also been used as an input for diagnostic prediction using machine learning methods which is the focus of this paper \cite{zhang2016diagpred}.

Identifying causal relationship \cite{spirtes2016causal} among diagnostic labels is extremely useful. The diagnostic labels in Figure \ref{fig:discomfort_drawing} includes symptom diagnoses, pattern diagnoses and psychophysiological diagnoses. The labels describe the symptoms, while pattern diagnosis and psychophysiological diagnosis explain the underlying cause of the symptoms. However, these labels are presented in an independent way since it is just the snapshot of a single patient. With access to a large number of cases, the causal relationship among these labels could be identified, providing a medical knowledge-graph of the relationship between symptoms and causes. Firstly, this causal graph by itself would be valuable for the decision making and education of health care personnel -- e.g., a clinician can get support in interpreting a pain drawing by using the causal graph as a template to look for likely causes of the observed symptoms. Secondly, the causal graph can be employed to make better automatic diagnostic prediction from pain drawings -- e.g., by using the identified relationships to refine the prediction results from other machine learning models \cite{zhang2016diagpred}. 

Our work explores utilizing causal identification algorithms to refine the diagnostic prediction and interpretation using discomfort drawings. Our contribution is two-fold:
\vspace{-1mm}
\begin{itemize}
\item We employ a causal identification algorithm, the Peter-Clark (PC) algorithm \cite{spirtes2001,spirtes2010}, to determine the causal relationship among diagnostic labels. 
\item We add a post-processing step to a diagnostic prediction algorithm, the Inter-Battery Topic Model (IBTM) \cite{zhang16IBTM}, using the identified causal graph to enhance both the detection accuracy and the interpretability of the result. 
\end{itemize}
\vspace{-1mm}
We present our method in Section \ref{sec:method},  experiments in Section \ref{sec:exp}, and discuss future work in Section \ref{sec:discussion}.

%% file: model.tex
\section{Method}
\label{sec:method}

\begin{figure}[b]
\centerline{\scalebox{0.35}{\input{model_figure.tex}}}
\caption{\footnotesize The method. The left panel shows the IBTM which is a multi-view latent space model that predict diagnostic labels given input. The middle panel demonstrates a causal graph which is learned from training data. The prediction is then used as the input for this causal graph to refine the result. The last panel demonstrates the final output, which is a subgraph instead of independent labels.}
\label{fig:method}
\end{figure}
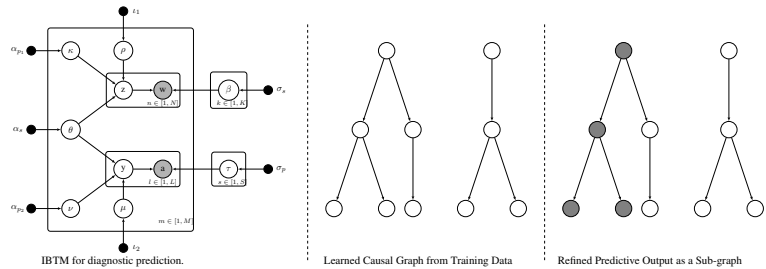

Our method of refining diagnostic prediction is presented in this section. The method is summarized in Figure \ref{fig:method}. The baseline model is IBTM (left panel) \cite{zhang16IBTM} as used in \cite{zhang2016diagpred}. IBTM is a latent variable model that learns a joint embedding $\theta$ of the discomfort drawing  $w$ and the diagnostic label $a$. Thus, it can be used for diagnostic prediction given an unseen discomfort drawing. Using training data, we identify a causal graph demonstrated in the middle panel. We present the causal identification method in Section \ref{sec:causalgraphs} with results from the data. At test time, we predict the diagnostic labels for an unseen pain drawing with IBTM. We then refine the predicted labels in the causal graph and obtain the final structured output as in the right panel in Figure \ref{fig:method}. We present this procedure in Section \ref{sec:refine_result}.

\vspace{-1mm}
\subsection{Causal Identification}
\label{sec:causalgraphs}
\vspace{-1mm}

\begin{figure}[t]
\centerline{\includegraphics[width=0.9\textwidth]{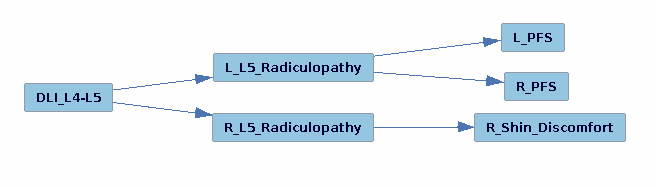}}
\vspace{-9mm}
\caption{\footnotesize Subgraph of diagnostic labels from Tetrad displaying the causal structure between DLI L4-L5, L5 radiculopathy, and patellofemoral pain syndrome (PFS) and right shinbone discomfort. We do not show the full graph due to space limit. }
\label{fig:subgraph1}
\vspace{-2mm}
\end{figure}

The causal graph should be structured as a \textit{directed acyclic graph} (DAG), where the relationships between the three types of diagnostic labels can be identified.  The properties of a DAG is that all edges must be directed from one node to another and that the graph does not allow any cycles, meaning that there can not exist a path from one node that leads to itself \cite{koller2009}. With some other constraints on the DAG, e.g., that interventions on a cause variable does not change the generating process from this cause to its effect, the DAG has a causal interpretation and represents causal relations between the variables~\cite{pearl2000,spirtes2001}.

To learn a DAG from data, a constraint-based search algorithm such as the PC (Peter-Clark) algorithm \cite{spirtes2001,spirtes2010} can be used. This algorithm starts from an undirected graph and performs a sequence of independence and conditional independence tests, and based on this builds an equivalence class, which may contain more than one DAG that shares the same independence and conditional independence relations in the data. As long as the conditional independence facts are known for the data, the PC algorithm can efficiently output such a equivalence class for both discrete and continuous data \cite{tetrad2017,spirtes2010}.  

We make the following additional constraints in the causal search (in additional to the independence and conditional independence constaints discovered from data): the diagnostic labels are separated into three categories, \textit{causes} $C$ (the psychophysiological diagnostic labels), \textit{reasons} $R$ (the pattern diagnostic labels) and \textit{symptoms} $S$ (the symptom diagnostic labels), where $C$, $R$ and $S$ are sets of these diagnoses. We also assume that these categories are structured as $C \rightarrow R \rightarrow S$, so that $C$ has a direct cause on $R$ and that $R$ directly causes $S$. Edges are forbidden within each individual set. We found that with those additional constraints, the output of PC was a DAG, with all edges oriented. Figure \ref{fig:subgraph1} shows a subgraph from a DAG built in Tetrad \cite{tetrad2017} with samples of the available diagnostic labels. The results tells that a disco-ligament injury at segment L4-L5 has a direct effect on the L5 nerve radiculopathy on both sides of the body. An injury on the left side of the nerve causes patellofemoral pain syndrome (PFS) on both sides of the body, while an injury on the right side causes discomforts on the right shinbone. Such a result is consistent with the distribution and connectivity of the nerves in the human body, indicating that the algorithm correctly identified the causal relationship among diagnostic labels in the dataset.

\vspace{-1mm}
\subsection{Refine Predictive Results using Causal Graph}
\label{sec:refine_result}
\vspace{-1mm}

\begin{algorithm}[b]
\caption{Update MPDs $p(x)$ in causal graph $G$.}
\label{alg:algorithm1}
\begin{algorithmic}
\ForAll {nodes $x \in G$}
	\State $p^{(0)}(x) \gets p_{IBTM}(x)$
\EndFor
\For {$t=1, \dots, \tau$}
  \ForAll {nodes $x \in G$}
      \ForAll {neighbours($x$) = $x'$}
     	 \State  Compute $p^{(t)*}(x')$ using $x$ and conditional probabilities between them. 
          \State $p^{(t)}(x') \gets \epsilon \,p^{(t)*}(x')  + (1-\epsilon) \, p^{(t-1)}(x')$
      \EndFor
  \EndFor
\EndFor
\end{algorithmic}
\end{algorithm}

The probability of each diagnostic label is predicted using IBTM. The conditional probability from cause to effect is obtained from the causal identification step. Given the learned DAG, we can then straightforwardly estimate the conditional distribution of each variable conditioning on its parents (or direct causes) from the given data.

Suppose we have obtained the prediction produced by IBTM. IBTM predicts each of the outputs separately, ignoring the dependence relationships among them.  As a consequence, it does not make full use of the dependence information in the data and the result is to be improved. Given the good performance of IBTM in single-output prediction, however, to achieve better predictions, it is sensible to leverage 1) the predicted marginal distributions for all outputs by IBTM and 2) the relationships among the variables implied by the qualitative DAG structure and the quantitative conditional distributions.
In other words, we can further update the prediction from IBTM, $p_{IBTM}(x)$, with the causal relationships given by graph $G$. The  probability for node $x$ is denoted by $p(x)$ and these are initialized with the $p_{IBTM}(x)$. For each time step $t$, we iterate through all of the nodes and update its neighboring nodes $x'$ with learning rate $\epsilon$.
The procedure of post-refining the predictive result is summarized in Algorithm \ref{alg:algorithm1}.

%% file: model_figure.tex
\begin{tikzpicture}

\tikzstyle{surround} = [thick,draw=black,rounded corners=1mm]

\tikzstyle{scalarnode} = [circle, draw, fill=white!11,  
    text width=1.2em, text badly centered, inner sep=2.5pt]

\tikzstyle{scalarnodeCyan} = [circle, draw=cyan, fill=white!11,  
    text width=1.2em, text badly centered, inner sep=2.5pt]
\tikzstyle{discnode}=[rectangle,draw,fill=white!11,minimum size=0.9cm]

\tikzstyle{Vnode}=[circle, radius=1pt,draw,fill=black]
\tikzstyle{vectornode} = [circle, draw, fill=white!11,  
    text width=2.3em, text badly centered, inner sep=2pt]
\tikzstyle{state} = [rectangle, draw, text centered, fill=white, 
    text width=8em, text height=6.7em, rounded corners]

\tikzstyle{arrowline} = [draw,color=black, -latex]
\tikzstyle{carrowline} = [line width=2pt, draw,color=black, -latex]
\tikzstyle{line} = [draw]


\node [Vnode] at ( 0.5, 0) (alpha_s){};
\node [] at ( 0, 0) (){$\alpha_s$};
\node [scalarnode] at ( 2, 0 ) (theta) { $\theta$ };
\node [Vnode] at ( 0.5, 3) (alpha_p1){};
\node [] at ( 0, 3) (){$\alpha_{p_1}$};
\node [scalarnode] at ( 2 , 3 ) (kappa) { $\kappa$ };
\node [scalarnode] at ( 4, 1.5) (z) {z};
\node [scalarnode, fill=black!30] at ( 5.5, 1.5) (w) {w};

\node [scalarnode] at ( 4, 3) (rho) {$\rho$};
\node [Vnode] at ( 4, 4.5) (iota1) {};
\node [] at ( 4.5, 4.5) () {$\iota_1$};

\node [Vnode] at ( 0.5, -3) (alpha_p2){};
\node [] at ( 0, -3) (){$\alpha_{p_2}$};
\node [scalarnode] at ( 2 , -3 ) (nu) { $\nu$ };
\node [scalarnode] at ( 4, -1.5) (y) {y};
\node [scalarnode, fill=black!30] at ( 5.5, -1.5) (a) {a};

\node [scalarnode] at ( 4, -3) (mu) {$\mu$};
\node [Vnode] at ( 4, -4.5) (iota2) {};
\node [] at ( 4.5, -4.5) () {$\iota_2$};

\node [scalarnode] at ( 8, 1.5) (beta) {$\beta$};
\node [Vnode] at ( 9.5, 1.5) (sigma_s1) {};
\node [] at ( 10, 1.5) () {$\sigma_{s}$};

\node [scalarnode] at ( 8, -1.5) (tau) {$\tau$};
\node [Vnode] at ( 9.5, -1.5) (sigma_p2) {};
\node [] at ( 10, -1.5) () {$\sigma_{p}$};
\node[surround, inner sep = .3cm] (f_N) [fit = (z)(w) ] {};
\node[surround, inner sep = .3cm] (f_L) [fit = (y)(a) ] {};
\node[surround, inner sep = .5cm] (f_M) [fit = (f_N)(f_L)(theta)(rho)(mu) ] {};

\node[surround, inner sep = .3cm] (f_beta) [fit = (beta) ] {};
\node[surround, inner sep = .3cm] (f_tau) [fit = (tau) ] {};

\node [] at (6, -3.5) (M) {\scriptsize $m \in [1,M]$};
\node [] at (5.5, -2) (L) {\scriptsize $l \in [1,L]$};
\node [] at (5.5, 1) (N) {\scriptsize $n \in [1,N]$};
\node [] at (8.15, 1) () {\scriptsize $k \in [1,K]$};
\node [] at (8.15, -2) () {\scriptsize $s \in [1,S]$};

\path [arrowline] (alpha_s) to (theta); 
\path [arrowline] (alpha_p1) to (kappa); 
\path [arrowline] (theta) to (z); 
\path [arrowline] (kappa) to (z); 
\path [arrowline] (rho) to (z); 
\path [arrowline] (z) to (w); 
\path [arrowline] (iota1) to (rho);

\path [arrowline] (alpha_p2) to (nu); 
\path [arrowline] (theta) to (y); 
\path [arrowline] (nu) to (y); 
\path [arrowline] (mu) to (y); 
\path [arrowline] (y) to (a); 
\path [arrowline] (iota2) to (mu);

\path [arrowline] (beta) to (w); 
\path [arrowline] (sigma_s1) to (beta); 

\path [arrowline] (tau) to (a); 
\path [arrowline] (sigma_p2) to (tau); 

\node[] at (3.6, -5) () {\large IBTM for diagnostic prediction.};
\draw[dashed] (11, 4) -- (11,-5);


\node [scalarnode] at ( 14, 3) (P1) {};
\node [scalarnode] at ( 13, 0) (R1) {};
\node [scalarnode] at ( 15, 0) (R2) {};
\node [scalarnode] at ( 12, -3) (S1) {};
\node [scalarnode] at ( 14, -3) (S2) {};
\node [scalarnode] at ( 15, -3) (S3) {};
\path [arrowline] (P1) to (R1); 
\path [arrowline] (P1) to (R2); 
\path [arrowline] (R1) to (S1); 
\path [arrowline] (R1) to (S2); 
\path [arrowline] (R2) to (S3); 

\node [scalarnode] at ( 18, 3) (P2) {};
\node [scalarnode] at ( 18, 0) (R3) {};
\node [scalarnode] at ( 17, -3) (S4) {};
\node [scalarnode] at ( 19, -3) (S5) {};
\path [arrowline] (P2) to (R3); 
\path [arrowline] (R3) to (S4); 
\path [arrowline] (R3) to (S5);

\node[] at (15.2, -5) () {\large Learned Causal Graph from Training Data};

\draw[dashed] (20, 4) -- (20,-5);


\node [scalarnode, fill=black!50] at ( 14+9, 3) (eP1) {};
\node [scalarnode, fill=black!50] at ( 13+9, 0) (eR1) {};
\node [scalarnode] at ( 15+9, 0) (eR2) {};
\node [scalarnode, fill=black!50] at ( 12+9, -3) (eS1) {};
\node [scalarnode, fill=black!50] at ( 14+9, -3) (eS2) {};
\node [scalarnode] at ( 15+9, -3) (eS3) {};
\path [arrowline] (eP1) to (eR1); 
\path [arrowline] (eP1) to (eR2); 
\path [arrowline] (eR1) to (eS1); 
\path [arrowline] (eR1) to (eS2); 
\path [arrowline] (eR2) to (eS3); 

\node [scalarnode] at ( 18+9, 3) (eP2) {};
\node [scalarnode] at ( 18+9, 0) (eR3) {};
\node [scalarnode] at ( 17+9, -3) (eS4) {};
\node [scalarnode] at ( 19+9, -3) (eS5) {};
\path [arrowline] (eP2) to (eR3); 
\path [arrowline] (eR3) to (eS4); 
\path [arrowline] (eR3) to (eS5);
\node[] at (24, -5) () {\large Refined Predictive Output as a Sub-graph};
\end{tikzpicture}

%% file: exp.tex
\section{Experiments}
\label{sec:exp}

\begin{figure}[t]
\centering 
\begin{minipage}{0.4\textwidth}
\includegraphics[height=3.8cm, width=\textwidth]{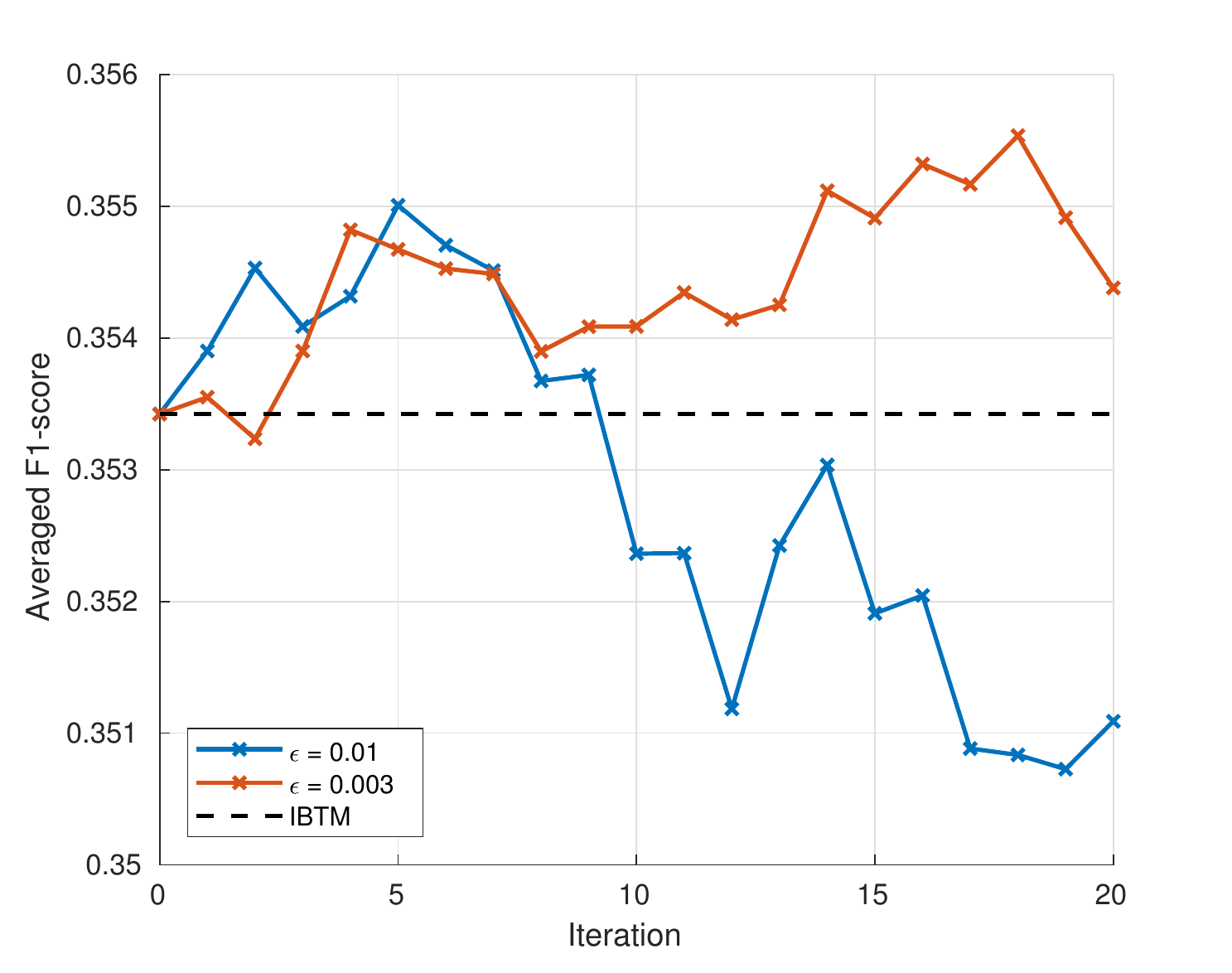}
\end{minipage}
\begin{minipage}{0.4\textwidth}
\includegraphics[height=3.8cm, width=\textwidth]{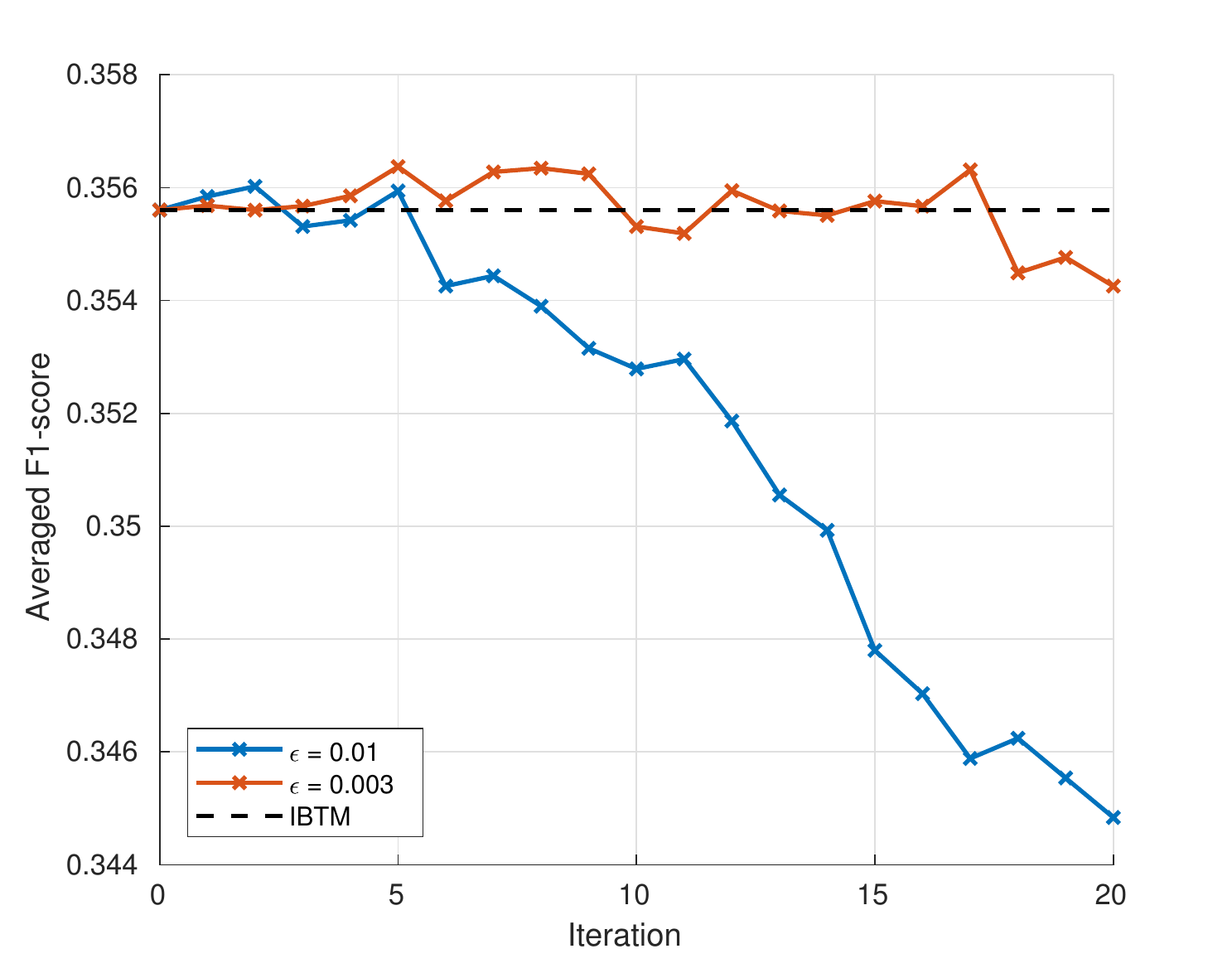}
\end{minipage}
\vspace{-2mm}
\caption{\footnotesize F-measures averaged over five random data splits with respect to 20 iterations of updating the causal graph, where IBTM was trained with $K=30$ (left) and $K=50$ (right). Two different update rates $\epsilon$ are used, and the black dashed line is the averaged results from IBTM.}
\label{fig:Fmeasures}
\end{figure}

\begin{table}[b]
\vspace{-2mm}
\caption{\footnotesize Example of unseen discomfort drawing (left), predicted diagnostic labels (middle) and resulting F-measures (right). Predictions made by IBTM are followed after \textbf{Prd IBTM}, the refined predictions are after \textbf{Prd DAG} and ground truth labels after \textbf{GT}. Successfully predicted labels are marked in blue, otherwise in red. 
}
\centering
\begin{tabular}{| c | p{7.1cm} | c |}
	\hline    
    \begin{minipage}{0.3\textwidth}
    	\includegraphics[width=\linewidth]{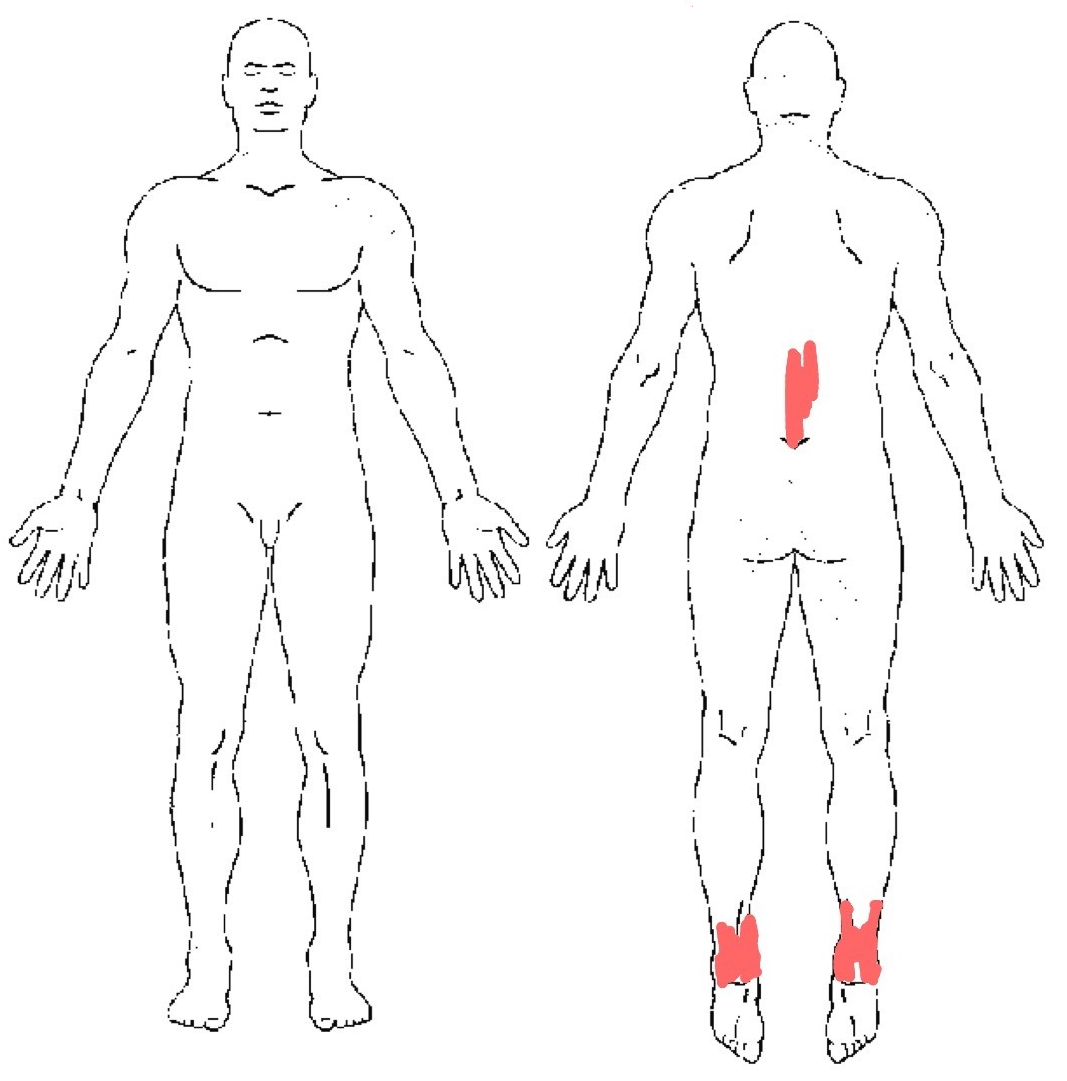}
    \end{minipage}
	&
    \begin{minipage}{0.6\textwidth}
    	\vspace{0.5em}
    	\small{
          \textbf{5 Prd IBTM:} \textcolor{blue}{L Calcaneal pain;}
          \\ \\
          \textcolor{blue}{DLI L5-S1; L S1 Rdc; R S1 Rdc;} \textcolor{red}{L C2 Rdc;}
          \\ \\
          \textbf{5 Prd DAG:} \textcolor{blue}{L Calcaneal pain; R Calcaneal pain*;}  
          \\ \\
          \textcolor{blue}{DLI L5-S1; R S1 Rdc; L S1 Rdc;}
          \\ \\
          \textbf{6 GT:}  \textcolor{red}{Lumbago;} \textcolor{blue}{R Calcaneal pain*; L Calcaneal pain;} 
          \\ \\
          \textcolor{blue}{DLI L5-S1; R S1 Rdc; L S1 Rdc; }
        }
        \vspace{0.5em}
    \end{minipage}
    &
    \begin{minipage}{0.1\textwidth}
    	\underline{\textbf{F1 IBTM}} \\
        72.73\%
        \\ \\
        \underline{\textbf{F1 DAG}} \\
        90.91\%
        \\ \\
        \underline{\textbf{Diff.}} \\
        18.18\%
    \end{minipage}
    \\ \hline
\end{tabular}
\label{tab:results_pred_gt}
\end{table}

We present our preliminary experimental results on refining diagnostic prediction. We use the same pain drawing dataset and parameter settings as in \cite{zhang2016diagpred}. The dataset consists of 174 pain drawings and 280 diagnostic labels, where 150 of them are symptom diagnoses ($S$), 61 are pattern diagnoses ($R$) and 69 are pyschophysiological diagnoses ($C$). We split the dataset randomly in two halves for training and testing. As in \cite{zhang2016diagpred}, the number of diagnostic labels to predict is determined by mean shift clustering for every test drawing. 
 
We run the updating procedure for 20 time steps and average the F-measure over all test examples on each iteration. We demonstrate the results with update rate $\epsilon = 0.01$ and $\epsilon = 0.003$ in Figure \ref{fig:Fmeasures}. IBTM was trained with the number of topics $K=30$ (left) and $K=50$ (right). The results show an increasing trend in the beginning iterations when $\epsilon = 0.003$ and $\epsilon = 0.01$ when $K=30$. 

Table \ref{tab:results_pred_gt} shows an example where the causal graph post-processing enhances the IBTM results. We see that R Calcaneal pain was not reported by IBTM because the probability was not high enough. On the other hand, with causal refining, the probability of Right Calcaneal pain increases, because it is highly probable when S1 Rdc is diagnosed. With our method, the result can be visualized to the clinician as a subgraph in the causal graph (omitted here due to space limitation).

%% file: discussion.tex
\section{Discussion}
\label{sec:discussion}

In this work, we employ a causal identification method, PC, to determine the causal relationship among various diagnostic labels. The identified causal graph is highly interpretable and is used to refine results from a diagnostic prediction model, IBTM.  

Our future work extends in two directions. Firstly, we will explore ways to visualize such causal graphs to medical students as an education tool, and to visualize subgraphs to clinicians, to enhance interpretability of automatic diagnostic results. Secondly, while our current automatic prediction method employs causal reasoning in a post-processing manner where IBTM and a casual graph are learned separately, we will in future work focus on structured prediction where casual reasoning is integrated into the machine learning model in an end-to-end training manner.